\documentclass{article}
\usepackage{spconf,amsmath,graphicx}

\usepackage{subcaption}
\usepackage{hyperref}

\title{Opening Deep Neural Networks with Generative Models}
\name{Marcos Vendramini$^{\dagger}$ \qquad Hugo Oliveira$^{\star}$ \qquad Alexei Machado$^{\ddag\P}$ \qquad Jefersson A. dos~Santos$^{\dagger}$}
\address{$^\dagger$ Department of Computer Science, Universidade Federal de Minas Gerais, Brazil \\ $^{\star}$ Institute of Mathematics and Statistics, University of S\~{a}o Paulo, Brazil \\ $^{\ddag}$ Department of Anatomy and Imaging, Universidade Federal de Minas Gerais, Brazil\\ $^{\P}$ Department of Computer Science, Pontifícia Universidade Católica de Minas Gerais, Brazil}
\usepackage[table,xcdraw]{xcolor}

\usepackage{multirow}

\begin{document}
%
\maketitle
\begin{abstract}
Image classification methods are usually trained to perform predictions taking into account a predefined group of known classes. Real-world problems, however, may not allow for a full knowledge of the input and label spaces, making failures in recognition a hazard to deep visual learning. Open set recognition methods are characterized by the ability to correctly identify inputs of known and unknown classes. In this context, we propose GeMOS: simple and plug-and-play open set recognition modules that can be attached to pre-trained Deep Neural Networks for visual recognition. The GeMOS framework pairs pre-trained Convolutional Neural Networks with generative models for open set recognition to extract open set scores for each sample, allowing for failure recognition in object recognition tasks. We conduct a thorough evaluation of the proposed method in comparison with state-of-the-art open set algorithms, finding that GeMOS either outperforms or is statistically indistinguishable from more complex and costly models.
\end{abstract}

\newcommand{\currprop}{\textwidth}

\section{Introduction}
\label{sec:introduction}


Image classification is typically modeled as a discriminative supervised learning problem. An image classifier is trained from a given set of images associated with a known number of classes, characterizing a {\em closed set} scenario~\cite{Geng:2020}. The model is expected to be effective while assigning a new image to the correct class. However, it is not capable of correctly identifying that an out-of-distribution (OOD) image belongs to an unknown class.

\begin{figure}
    \centering
    \begin{subfigure}[b]{0.15\textwidth}
        \centering
        \includegraphics[width=\textwidth]{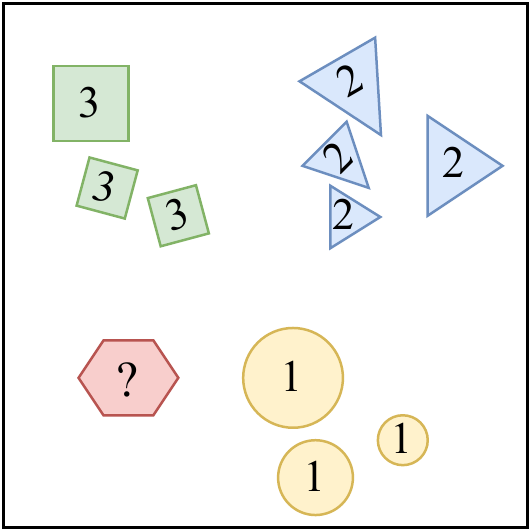}
        \caption{Dataset.}
        \label{fig:osr_definition_classes}
    \end{subfigure}
    \hfill
    \begin{subfigure}[b]{0.15\textwidth}
        \centering
        \includegraphics[width=\textwidth]{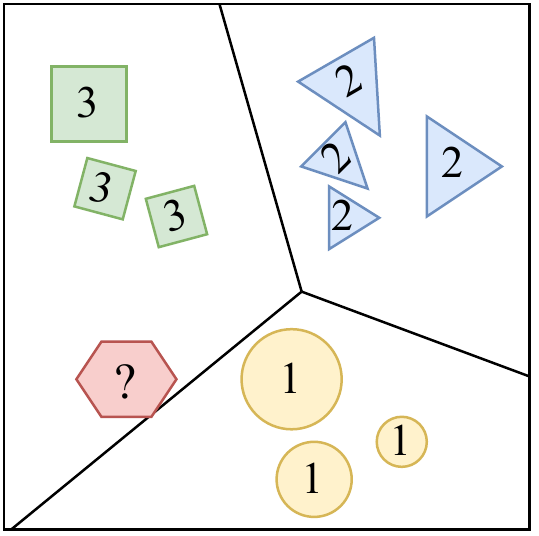}
        \caption{Closed set space.}
        \label{fig:osr_definition_closed}
    \end{subfigure}
    \hfill
    \begin{subfigure}[b]{0.15\textwidth}
        \centering
        \includegraphics[width=\textwidth]{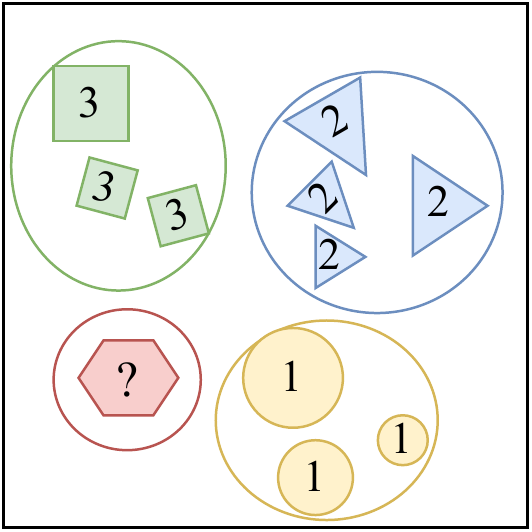}
        \caption{Open set space.}
        \label{fig:osr_definition_open}
    \end{subfigure}
    \caption{Example of an OSR scenario. Given a data set (a) with three known classes (\textit{1}, \textit{2} and \textit{3}) and one unknown class (``?''), closed set methods partition the space considering only known classes (b) while in open set methods each known class is assigned a limited region in the feature space.}
    \label{fig:osr_definition}
\end{figure}

On the other hand, {\em Open Set Recognition} (OSR) is characterized by the ability of correctly classifying inputs of known and unknown classes. The main challenge of OSR is to correctly delineate a decision boundary between potentially similar known and unknown classes without ever having access to samples from the unknowns. According to Scheirer \textit{et al.}~\cite{Scheirer:2012}, during the inference phase, an OSR system should be able to correctly classify the instances/pixels of the classes used for training (Known Known Classes -- KKCs) while recognizing the samples of classes that were \textbf{not} seen during training (Unknown Unknown Classes -- UUCs). The differences between open and closed set scenarios are exemplified in Figure \ref{fig:osr_definition}.

Based on this definition, the main difference between closed and open set scenarios can be associated to the degree of knowledge of the world, e.g., the awareness of all possible classes. Specifically, while in the closed set scenario the methods should have full knowledge of the world, open set approaches must assume that they do not have access to all possible classes during training. 
As deep visual learning assumes full knowledge of class space, traditional implementations of Convolutional Neural Networks (CNNs) \cite{krizhevsky2009learning} are inherently closed set, rendering them unsuitable for detecting recognition failures on their own.

Aiming to adapt Deep Neural Networks (DNNs) for open set scenarios, in this paper we propose a novel plug-and-play method for opening pretrained closed set CNNs. We propose the \textbf{Ge}nerative \textbf{M}odels for \textbf{O}pen \textbf{S}et recognition (GeMOS) framework for opening pretrained CNNs. In order to avoid the costly training of additional deep models as several of the baselines do (Section~\ref{sec:related}), we use simpler shallow generative modeling to compute the likelihoods of each sample pertaining or not to the data distribution (Section~\ref{sec:method}). We conduct thorough evaluations of the performance of multiple generative models among themselves and in comparison to state-of-the-art baselines (Sections~\ref{sec:setup} and~\ref{sec:results}). At last, we show our conclusions about the method in Section~\ref{sec:conclusion}.




\section{Related Work}
\label{sec:related}





The concept of OSR was introduced by Scheirer \textit{et al.} \cite{Scheirer:2012} and their proposed approach, as well as other early studies, were based on shallow decision models such as threshold-based or support-vector-based methods.

Recent trends in the literature coupled OSR's failure recognition capabilities with the versatility and modeling power of Deep Learning. Early strategies for deep OSR~\cite{Bendale:2016,Ge:2017,Liang:2017} consisted of incorporating the UUCs detection directly into the prediction of the DNN. For instance, OpenMax~\cite{Bendale:2016} performed OSR by reweighting SoftMax activation probabilities to account for a UUC during test time. Many improvements to OpenMax were further proposed such as the association of Generative Adversarial Networks (GANs) \cite{goodfellow2014generative} to provide synthetic images to the method (G-OpenMax~\cite{Ge:2017}). Out-of-Distribution Detector for Neural Networks (ODIN)~\cite{Liang:2017} inserted small perturbations in the input image in order to increase the separability in the SoftMax predictions between in- and out-of-distribution data. This strategy allowed ODIN to work similarly to OpenMax~\cite{Bendale:2016} and to operate close to the label space, using a threshold over class probabilities to discern between KKCs and UUCs.

More recently, OSR for deep image classification have incorporated input reconstruction error in supervised DNN training as a way to identify OOD samples~\cite{Yoshihashi:2019,Oza:2019,Sun:2020} by employing generative strategies. Classification-Reconstruction learning for Open-Set Recognition (CROSR)~\cite{Yoshihashi:2019} jointly trains a supervised DNN for classification and an AutoEncoder (AE) to encode the input into an bottleneck embedding and then decodes it to reconstruct. The magnitude of the reconstruction error can then be used to delineate between known and unknown classes. Class Conditional AutoEncoder (C2AE)~\cite{Oza:2019}, similarly to CROSR, uses the reconstruction error of the input from an AE and EVT modeling to determine a threshold that is able to discern between KKC and UUC samples. Conditional Gaussian Distribution Learning (CGDL)~\cite{Sun:2020} uses a Variational AutoEncoder (VAE) to model the bottleneck representation of the input images and their activations according to a vector of gaussian means and standard deviations in a lower-dimensional high semantic-level space. From this multivariate gaussian model, Kullback–Leibler (KL) divergences can be computed for each sample and thresholded such that 95\% of the training examples of a class remain within the cluster of this KKC.

\section{Proposed Method}
\label{sec:method}


This section details GeMOS, a novel approach composed of a CNN that extracts feature-level information from KKCs, and of generative models that use these features to assign a score for each KKC sample and identify UUCs.

GeMOS uses an off-the-shelf pretrained CNN for image classification as backbone in a closed set scenario. This backbone is used to extract feature-level information from the intermediary activations (e.g. $a^{(L_3)}$, $a^{(L_4)}$, $a^{(L_5)}$, $a^{(L_6)}$ from Figure~\ref{fig:gemos_architecture}) to feed the generative models. Convolutional layer activations are linearized using a function $\phi$ -- usually an average pooling -- and concatenated into a single vector ($a^{\star}$) for each sample. $a^{\star}$ can then be fed to the generative model $G$, which learns the most representative feature maps to maximize the reconstruction of this feature vector. The inherent dimensionality reduction of $G$ increases the likelihood that only the most relevant activations are taken into account to compute sample scores.

\renewcommand{\currprop}{0.85\textwidth}
\begin{figure*}
    \centering
    \includegraphics[width=\currprop]{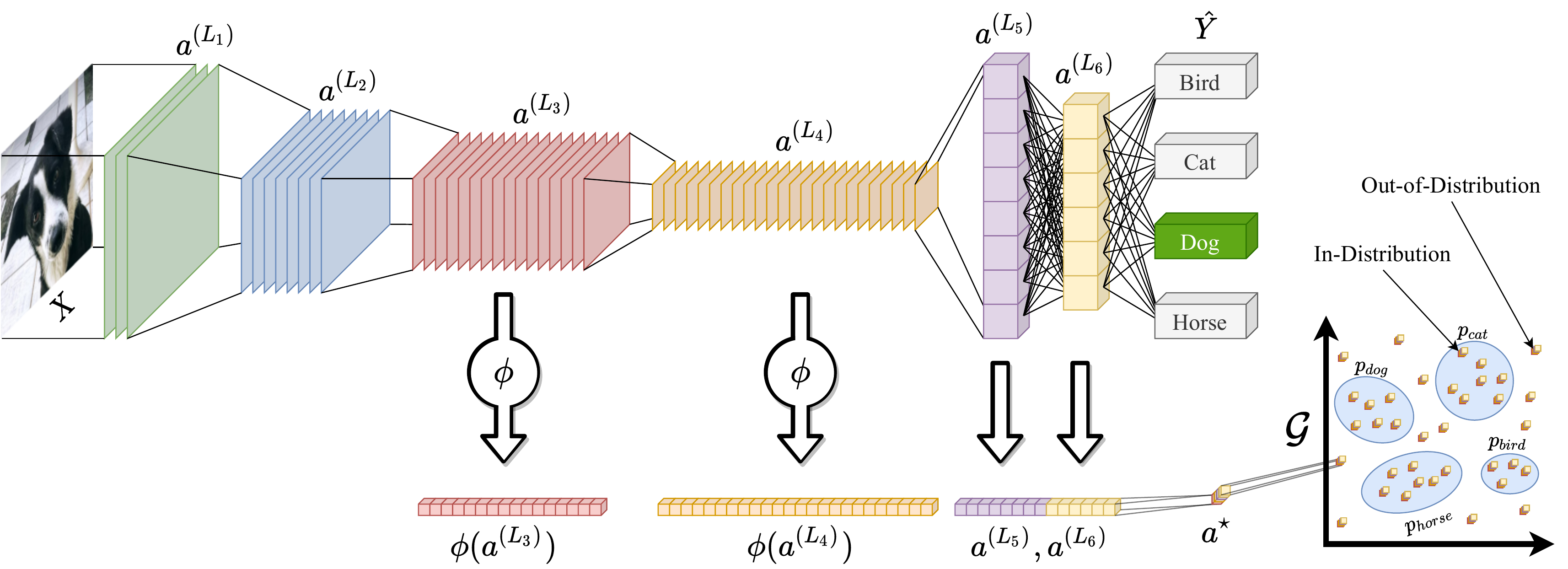}
    \caption{Depiction of the GeMOS pipeline. A CNN model is used to classify and extract the intermediary activations $a^{(L_3)}$, $a^{(L_4)}$, $a^{(L_5)}$ and $a^{(L_6)}$. In this example, $a^{(L_3)}$ and $a^{(L_4)}$ are 2D activation maps from convolutional layers, while $a^{(L_5)}$ and $a^{(L_6)}$ are originally 1D. In order to concatenate these features, we pool $a^{(L_3)}$ and $a^{(L_4)}$ into a 1D vector using a function $\phi$ (e.g. average pooling) and concatenate them with $a^{(L_5)}$ and $a^{(L_6)}$, resulting in a feature vector $a^{\star}$. $a^{\star}$ can then be used as input to $G$.}
    \label{fig:gemos_architecture}
\end{figure*}

In order to solve the OSR task, GeMOS employs a set $\mathcal{G} = \{k_{0}, k_{1}, \dots, k_{C-1}\}$ of generative models, with $C$ representing the number of KKCs. During the training process, each generative model $k_{c}$ is trained with the activation values ($a^{\star}$) of samples correctly predicted to be from the corresponding class $c$ in the training set. Multiple generative models were tested, such as Principal Component Analysis (PCA) \cite{tipping1999mixtures-pca}, One-Class Support Vector Machine (OCSVM) \cite{scholkopf2001estimating-svm}, Isolation Forest (IF) \cite{liu2008isolation} and Local Outlier Factor (LOF) \cite{breunig2000lof}. All of these traditional methods for dimensionality reduction and/or OOD detection assume unimodal distributions within each class, thus we also inserted in our experiments Gaussian Mixture Models (GMM) \cite{bishop2006pattern-gmm} in order to test a multimodal strategy. One should notice, however, that any generative model that produces a likelihood score for its samples can be used. Source code for GeMOS is available on GitHub\footnotemark\footnotetext{\url{https://github.com/marcosvendramini/GeMOS}}.

\noindent\textbf{OOD Scoring:} Closed set image recognition is performed by the CNN, while OSR is introduced into the framework by thresholding on scores returned by $\mathcal{G}$. Each $k_{c} \in \mathcal{G}$ fits one class and can generate a likelihood score $l_{i,c} \approx p_{c}$ for an input $x_{i}$ to express how similar the corresponding vector $a^{\star}_{i}$ is to the probability distribution $p_{c}$ of class $c$, predicted by the CNN. In order to identify if $x_{i}$ belongs to an unknown class, a threshold is applied to $l_{i,c}$ to define if the input is in-distribution or out-of-distribution. We do not fix the threshold to one single cutoff value, as it depends on the hyperparameters of the generative model, its scoring function and the failure tolerance of the application. This is due to the fact that there is always a trade-off between KKC accuracy and UUC identification performance. Some scenarios might require minimal interference in the performance of the KKCs (e.g. medical image analysis), while others can tolerate a higher penalty in the performance of known classes to more reliably identify OOD samples (e.g. anomaly detection in videos).

A main advantage of GeMOS in comparison to end-to-end trainable deep models is the fact that shallow generative models can be attached to any pretrained deep closed set model with minimal additional computation time. Contrarily to other state-of-the-art methods that rely on AE reconstruction error \cite{Yoshihashi:2019,Oza:2019}, GeMOS focuses on generative models that do not require GPU for training, rendering it an ideal candidate for a plug-and-play alternative for opening pretrained CNNs.



\section{Experimental Setup}
\label{sec:setup}

\noindent\textbf{Datasets:} Following common protocol on OSR literature \cite{Yoshihashi:2019,Oza:2019,Sun:2020}, two image sets were used as in-distribution (KKCs) data: MNIST \cite{lecun1998gradient} and CIFAR-10 (CF10) \cite{krizhevsky2009learning}. Both datasets contain 10 classes with annotations for each sample and thousands of examples per class. For $\mathcal{D}^{\text{in}}=\text{MNIST}$, we set as $\mathcal{D}^{\text{out}}$ other handwritten character recognition datasets, as these are related but OOD data. More specifically, we use Omniglot \cite{lake2015human}, EMNIST \cite{cohen2017emnist} and KMNIST \cite{clanuwat2018deep} as $\mathcal{D}^{\text{out}}$ for MNIST. Concerning the CF10 dataset we have used CIFAR-100 (CF100) \cite{krizhevsky2009learning}, Tiny-ImageNet (TIN) \cite{le2015tiny} and a commonly used subset LSUN \cite{yu2015lsun} as $\mathcal{D}^{\text{out}}$. 


\noindent\textbf{Backbones and Computational Resources:} As the proposed method is intended to be plug-and-play on existing DNNs from the start, some of the most common CNN backbones for visual recognition were considered. More specifically, for MNIST we used the ResNet-18 \cite{he2016deep} -- as exploratory experiments showed that more complex backbones did not result in a noticeable gain in performance for this dataset -- and for CF10 the WideResNet (WRN-28-10) \cite{zagoruyko2016wide} and DenseNet-121 \cite{huang2017densely} architectures were evaluated. In order to pretrain these CNNs on $\mathcal{D}^{\text{in}}$, standard procedures were followed for achieving state-of-the-art performance in these datasets: training for 400 epochs with SGD, learning rate of $2 \times 10^{-2}$ scheduled to decrease by a factor of 10 in epochs 200 and 300, L2 regularization of $5 \times 10^{-4}$ and online data augmentation (e.g. random horizontal flips and random crops). Batch size was set to 32 samples, as we wanted to limit all experiments to a single GPU with 12GB of memory. All DNNs were implemented using the Pytorch\footnotemark\footnotetext{\url{https://pytorch.org/}} framework, while scikit-learn\footnotemark\footnotetext{\url{https://scikit-learn.org/}} was used for the generative models.


\noindent\textbf{Metrics:} Again following experimental procedures from the literature \cite{Yoshihashi:2019}, we use a set of threshold-dependent and threshold-independent metrics to assess the performance of GeMOS over both KKCs and UUCs. The binary Area Under Curve (AUC) in the classification task between KKCs and UUCs was the main threshold-independent metric, capturing information about the performance of the model without requiring a cutoff point that separates in- from out-of-distribution samples. Our main threshold-dependent metric is the F1-score, commonly used to assess the performance of models on supervised tasks. In this case, we group all UUCs in one single class and compute the F1-score accordingly for all KKCs and the sole UUC. 
As F1 is a threshold-dependent metric and we do not fix a single threshold value, we use the 5-fold cross-validation to determine the best cut-off and report this F1 value on Section~\ref{sec:results}.

\section{Results and Discussion}
\label{sec:results}

\subsection{Unimodal vs. Multimodal Generative Models}
\label{sec:ablation}

\begin{table*}[!t]
    \footnotesize
    \centering
    \caption{AUC and F1-score comparison for different generative models on $\mathcal{D}^{\text{out}}= \text{CF10}/\mathcal{D}^{\text{out}}= \text{CF100}$.}
    \label{tab:cifar_ablation}
    \begin{tabular}{cccccccccccc}
        \hline
        \textbf{Metrics} & \textbf{GMM2} & \textbf{GMM4} & \textbf{GMM8}          & \textbf{GMM16}         & \textbf{PCA2} & \textbf{PCA4} & \textbf{PCA8} & \textbf{PCA16} & \textbf{IF}   & \textbf{LOF}  & \textbf{OCSVM} \\ \hline
        \textbf{F1}      & 0.83 & 0.83 & \textbf{0.84} & \textbf{0.84} & 0.82 & 0.82 & 0.83 & 0.83  & 0.81 & 0.83 & 0.82  \\ 
        \textbf{AUC}     & 0.90 & 0.91 & \textbf{0.91} & \textbf{0.91} & 0.89 & 0.89 & 0.90 & 0.90  & 0.89 & 0.90 & 0.89  \\ \hline
    \end{tabular}
\end{table*}

Table~\ref{tab:cifar_ablation} shows F1 and AUC scores for GMM with 2, 4, 8 and 16 components, PCA with 2, 4, 8 and 16 components, IF, LOF and OCSVM using DenseNet as backbone with $\mathcal{D}^{\text{in}}=\text{CF10}$ and $\mathcal{D}^{\text{out}}=\text{CF100}$. GMM with 8 and 16 components presented the best results and, thus, GMM8 was the model chosen as standard for the baseline comparisons in Section~\ref{sec:inter_dataset}. The optimal number of components on GMM may vary with dataset and task, for our tested cases, GMM8 produce better scores compared to the other number of components.

We attribute the good performance of GMM to its multimodality capabilities, which is not commonly found in reconstruction-based DNNs as AEs or VAEs or in other shallow generative models (e.g. PCA, IF, LOF and OCSVM). This is due to the fact that real-world data, even within a single class, does not commonly follow unimodal distributions. 

\subsection{Baseline Comparison}
\label{sec:inter_dataset}

\begin{table}[!t]
    \footnotesize
    \centering
    \caption{F1-score and AUC for OSR with $\mathcal{D}^{\text{in}}=\text{MNIST}$.}
    \label{tab:result_din_mnist}
    \begin{tabular}{ccccccc}
        \hline
        \textbf{$\mathcal{D}^{\text{out}}$}       & \multicolumn{2}{c}{\textbf{Omniglot}} & \multicolumn{2}{c}{\textbf{EMNIST}} & \multicolumn{2}{c}{\textbf{KMNIST}} \\ 
        \textbf{Metrics}        & \textbf{F1}        & \textbf{AUC}      & \textbf{F1}       & \textbf{AUC}     & \textbf{F1}       & \textbf{AUC}     \\ \hline
        \textbf{SoftMax$^\tau$~\cite{Sun:2020}} & 0.60               & --                & --                & --               & --                & --               \\ 
        \textbf{OpenMax~\cite{Bendale:2016}}        & 0.78               & --                & --                & --               & --                & --               \\ 
        \textbf{ODIN~\cite{Liang:2017}}           & --                 & \textbf{1.00}     & --                & --               & --                & --               \\ 
        \textbf{CROSR~\cite{Yoshihashi:2019}}          & 0.79               & --                & --                & --               & --                & --               \\         \textbf{C2AE~\cite{Oza:2019}}           & --                 & --                & --                & --               & --                & --               \\ 
        \textbf{CGDL~\cite{Sun:2020}}           & 0.85               & --                & --                & --               & --                & --               \\ 
        \textbf{GeMOS}          & \textbf{0.91}      & 0.97              & \textbf{0.74}     & \textbf{0.97}    & \textbf{0.92}     & \textbf{0.99}    \\ \hline
    \end{tabular}
\end{table}

\begin{table}[!t]
    \footnotesize
    \centering
    \caption{F1-score for OSR with $\mathcal{D}^{\text{in}}=\text{CF10}$.}
    \label{tab:result_din_cifar_f1}
    \begin{tabular}{cccccc}
        \hline
        \textbf{$\mathcal{D}^{\text{out}}$} & \textbf{CF100} & \textbf{\begin{tabular}[c]{@{}c@{}}TIN\\ (crop)\end{tabular}} & \textbf{\begin{tabular}[c]{@{}c@{}}TIN\\ (resize)\end{tabular}} & \textbf{\begin{tabular}[c]{@{}c@{}}LSUN\\ (crop)\end{tabular}} & \textbf{\begin{tabular}[c]{@{}c@{}}LSUN\\ (resize)\end{tabular}} \\ \hline
        \textbf{CROSR~\cite{Yoshihashi:2019}}    & --                & 0.72                                                          & 0.74                                                            & 0.72                                                           & 0.75                                                             \\ 
        \textbf{C2AE~\cite{Oza:2019}}     & --                & 0.84                                                          & 0.83                                                            & 0.78                                                           & 0.80                                                             \\ 
        \textbf{CGDL~\cite{Sun:2020}}     & --                & 0.84                                                          & 0.83                                                            & 0.81                                                           & 0.81                                                             \\ 
        \textbf{GeMOS}    & \textbf{0.80}     & \textbf{0.92}                                                 & \textbf{0.92}                                                   & \textbf{0.90}                                                  & \textbf{0.93}
        \\ \hline
    \end{tabular}
\end{table}

\begin{table}[!t]
    \footnotesize
    \centering
    \caption{AUC for OSR with $\mathcal{D}^{\text{in}}=\text{CF10}$.}
    \label{tab:result_din_cifar_auc}
    \begin{tabular}{cccccc}
        \hline
        \textbf{$\mathcal{D}^{\text{out}}$} & \textbf{CF100} & \textbf{\begin{tabular}[c]{@{}c@{}}TIN\\ (crop)\end{tabular}} & \textbf{\begin{tabular}[c]{@{}c@{}}TIN\\ (resize)\end{tabular}} & \textbf{\begin{tabular}[c]{@{}c@{}}LSUN\\ (crop)\end{tabular}} & \textbf{\begin{tabular}[c]{@{}c@{}}LSUN\\ (resize)\end{tabular}} \\ \hline
        \textbf{ODIN~\cite{Liang:2017}}     & \textbf{0.90}     & 0.94                                                          & 0.92                                                            & 0.96                                                           & 0.95                                                             \\ 
        \textbf{OODNN~\cite{devries2018learning}}    & --                & 0.96                                                          & 0.95                                                            & \textbf{0.97}                                                  & 0.96                                                             \\ 
        \textbf{MSP~\cite{hendrycks2016baseline}}      & 0.88              & --                                                            & --                                                              & --                                                             & --                                                               \\ 
        \textbf{GeMOS}    & 0.88              & \textbf{0.98}                                                 & \textbf{0.99}                                                   & \textbf{0.97}                                                  & \textbf{0.99}                                                    \\ \hline
    \end{tabular}
\end{table}

Table~\ref{tab:result_din_mnist} shows the F1 and AUC scores for $\mathcal{D}^{\text{in}}=\text{MNIST}$ and Omniglot, EMNIST and KMNIST as $\mathcal{D}^{\text{out}}$. In this case, GeMOS was tested with ResNet-18 as backbone and GMM with 8 components. 
The results for 8 components are reported in Table \ref{tab:result_din_mnist}, with GeMOS presenting the best results in F1-score on Omniglot followed by CGDL, CRSOR, OpenMax and SoftMax thresholding. The other metrics and dataset were not reported in the literature, except by the AUC on Omniglot where ODIN outperform GeMOS. Comparing the results for every datasets, GeMOS shows to be consistently better than other state-of-art architectures.

Tables~\ref{tab:result_din_cifar_f1} and~\ref{tab:result_din_cifar_auc} show the F1 and AUC scores respectively for $\mathcal{D}^{\text{in}}=\text{CF10}$ and CF100, TIN (crop), TIN (resize), LSUN (crop), LSUN (resize) as $\mathcal{D}^{\text{out}}$. In this experiment, GeMOS was tested using WRN-28-10 as backbone and GMM with 8 components.
GeMOS outperforms all other methods in 4 out of 5 datasets, except for the AUC of OODNN on LSUN (crop), that achieved the same score. On CF100, ODIN achieved the best AUC followed by GeMOS and MSP. Due to the fact that CF10 is considerably more complex than MNIST, these results show consistency in the OOD detection capabilities of GeMOS even in harder scenarios. 


\section{Conclusion}
\label{sec:conclusion}

In this work, we propose GeMOS, a simple novel method for Open Set Recognition. GeMOS uses generative models to identify OOD samples based on feature vectors extracted from pre-trained closed set CNNs. The consistency of the proposed method was assessed by using different backbones and generative models tested in several datasets used as in- and out-of-distribution. Our ablation study showed that GMMs perform consistently better than unimodal shallow methods in OSR. The results were compared to state-of-the-art OSR methods mimicking their exact test protocol, with GeMOS consistently achieving the best results. GeMOS' performance can be attributed to the fact that we feed the intermediary feature vectors from the CNN to the generative model instead of looking only at the input \cite{Yoshihashi:2019,Oza:2019} or output spaces \cite{Bendale:2016}.

The main drawback of GeMOS observed in our exploratory experiments was the high reliance on the performance of the closed set model, where a dip of around 5\% in KKC accuracy severely compromised the OSR detection capabilities of the algorithm. However, given a state-of-the-art pre-trained CNN in the KKCs, GeMOS can yield equivalent results or even outperform more costly alternatives. Additionally, GeMOS has the advantage of being able to be coupled to any pre-trained CNN, while most reconstruction-based methods (e.g. C2AE and CGDL) require expensive training of additional neural network architectures.


\section*{Acknowledgments}
We are grateful to the Organization of Latin American and Caribbean Supreme Audit Institutions (OLACEFS), CAPES (Finance Code 001), CNPq (grants \#424700/2018-2 and \#311395/2018-0), and FAPEMIG (grant APQ-00449-17). We also thank NVIDIA for the donation of GPUs.

\bibliographystyle{IEEEbib}
\bibliography{refs}

\end{document}